\documentclass[sigconf]{acmart}

\usepackage{booktabs} 

\usepackage{graphicx}
\usepackage{amsmath}
\usepackage{algorithm} 
\usepackage{algorithmic} 
\usepackage{multirow}
\usepackage{amsmath}
\usepackage{caption}
\usepackage{hyperref}
\usepackage{arydshln}

\setcopyright{none}
\acmArticle{4}
\acmPrice{15.00}

\begin{document}
\title{Towards Time-Aware Distant Supervision\\ for Relation Extraction}


\author{Tianwen Jiang}
\affiliation{%
	\institution{Harbin Institute of Technology}
}
\email{twjiang@ir.hit.edu.cn}

\author{Sendong Zhao}
\affiliation{%
	\institution{Harbin Institute of Technology}
}
\email{sdzhao@ir.hit.edu.cn}

\author{Jing Liu}
\affiliation{%
	\institution{Microsoft Research Asia}
}
\email{liudani@microsoft.com}

\author{Jin-Ge Yao}
\affiliation{%
	\institution{Microsoft Research Asia}
}
\email{Jinge.Yao@microsoft.com}

\author{Ming Liu}
\affiliation{%
	\institution{Harbin Institute of Technology}
}
\email{mliu@ir.hit.edu.cn}

\author{Bing Qin}
\affiliation{%
	\institution{Harbin Institute of Technology}
}
\email{bqin@ir.hit.edu.cn}

\author{Ting Liu}
\affiliation{%
	\institution{Harbin Institute of Technology}
}
\email{tliu@ir.hit.edu.cn}

\author{Chin-Yew Lin}
\affiliation{%
	\institution{Microsoft Research Asia}
}
\email{cyl@microsoft.com}

\renewcommand{\shortauthors}{}

\begin{abstract}
Distant supervision for relation extraction heavily suffers from the wrong labeling problem. To alleviate this issue in news data with the timestamp, we take a new factor \textit{\textbf{time}} into consideration and propose a novel time-aware distant supervision framework (\textsc{Time-DS}). \textsc{Time-DS} is composed of a time series \textit{instance-popularity} and two strategies. \textit{Instance-popularity} is to encode the strong relevance of time and true relation mention. Therefore, \textit{instance-popularity} would be an effective clue to reduce the noises generated through distant supervision labeling. The two strategies, i.e., hard filter and curriculum learning are both ways to implement \textit{instance-popularity} for better relation extraction in the manner of \textsc{Time-DS}. The curriculum learning is a more sophisticated and flexible way to exploit \textit{instance-popularity} to eliminate the bad effects of noises, thus get better relation extraction performance. Experiments on our collected multi-source news corpus show that \textsc{Time-DS} achieves significant improvements for relation extraction.
\end{abstract}

%
%



\maketitle

\section{Introduction}
Distant supervision (DS) has become a popular paradigm for relation extraction in recent years~\cite{mintz2009distant,zeng2015distant,zheng2017joint}. Distant supervision could largely extend annotated training instances through aligning relation instances in knowledge bases (KB) to sentences in text.

However, distant supervision heavily suffers from the wrong labeling problem because the aligned sentences are not necessarily expressing the same relations as the ones in KB\cite{mintz2009distant,riedel2010modeling}. Such wrong labeling problem introduces many false positive training instances that hurts the performance of the models. Many efforts have been made to alleviate the bad effects of such noises produced by DS. Some studies~\cite{riedel2010modeling,surdeanu2012multi,ritter2013modeling,min2013distant} applied multi-instance learning for relaxing the distant supervision assumption and making the at-least-one assumption: if two entities preserve a relation in a KB, at least one sentence that mentions the entity pair expresses the relation. Nowadays, Some neural networks studies\cite{zeng2015distant,lin2016neural,feng2017effective} learned from multiple instances attentively, without explicitly characterizing the inherent noise. However, all these studies attempted to empower noise-tolerance of models rather than reducing the noises from the source, i.e., the process of distant supervision. Therefore, these studies still suffer from the effects of noises by some degree. 

\begin{figure}[tb]
	\centering 
	\includegraphics[width=8.5cm]{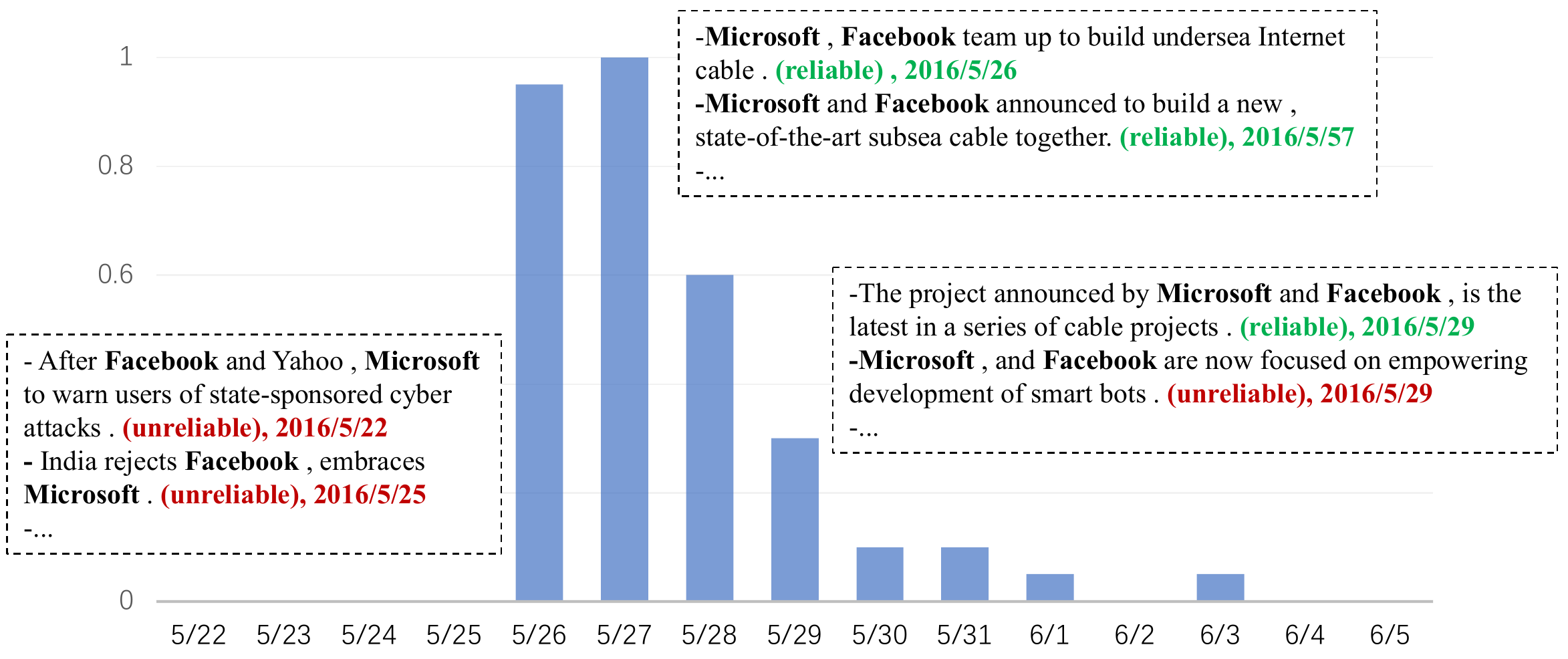}
    \vspace{-0.2in}
	\caption{An illustration of the factor \textit{\textbf{time}} in affecting the distribution of the relation \textsc{Partnership}(\textit{Microsoft, Facebook}) in news corpus of 2016. 20 aligned sentences for the relation instance from distant supervision are sampled randomly for each day, and each bar represents the ratio of sentences which really express the relation \textsc{Partnership}(\textit{Microsoft, Facebook}) in that day, i.e. the ratio of reliable sentences. The ratio is nearly zero before May 26, then a sudden increase occurred in May 26 and sustained for about two days. Gradual decrease is found after May 27.}
	\label{fig: insight}
    \vspace{-0.2in}
\end{figure}


Taking \textit{\textbf{time}} into consideration can effectively alleviate the noises in DS for relation extraction. We find that relation instances in news data is usually time-sensitive thus not uniformly distributed in news, e.g., \textsc{Partnership}(\textit{Microsoft, Facebook}) in Figure~\ref{fig: insight}. The mentions of \textsc{Partnership}(\textit{Microsoft, Facebook}) tend to be concentrated in a certain period of time, i.e. from May 26 to 27 in 2016. Therefore, an intuition is that the automatically annotated dataset produced in such a certain period of time has extremely fewer noises, while mentions in other days are more likely false positives. 

To model the above intuition that introducing a new factor \textit{\textbf{time}} to enhance DS for automatic dataset annotation, we propose a novel time-aware distant supervision framework (\textsc{Time-DS}). \textsc{Time-DS} can effectively reduce the impact of noises by making use of \textit{\textbf{time}}. \textsc{Time-DS} uses a time series \textit{instance-popularity} for each relation instance to indicate how many news mentioning the relation every day. The \textit{instance-popularity} is proposed to encode the strong relevance of time and true relation mention. For better use of the time series \textit{instance-popularity}, \textsc{Time-DS} considers two strategies. 

First, taking \textit{instance-popularity} as a hard filter to eliminate noises in the process of DS, i.e., aligning. This hard filter sets a hard threshold to filter noisy data, i.e. unreliably aligned sentences. This is a simple strategy with apparent drawbacks. It (1) heavily relies on the threshold of \textit{instance-popularity}, and (2) unable to utilize the noises in training to make DS-based models more robust. We therefore propose a second strategy to conduct curriculum learning~\cite{bengio2009curriculum} on the weighted training instances, i.e., instances with \textit{instance-popularity}, which is a slightly more sophisticated but flexible way to exploit \textit{instance-popularity}. The main idea of curriculum learning is simple: starting with the easiest aspect of a task, and leveling up the difficulty gradually. In this study, we begin with the high-quality annotated sentences and gradually add low-quality sentences into training set according to our proposed \textit{instance-popularity}. Curriculum learning for \textsc{Time-DS} can make full use of every weighted training instance for better relation extraction performance, while obtaining a robust model to put up with noises. 


We conduct experiments of relation extraction on a \textit{multi-source} news corpus with timestamp, where it is more natural to utilize rich temporal statistics compared with independent single documents only. Meanwhile, the \textit{multi-source} news corpus is of value for training a DS model for relation extraction from two other aspects. First, the \textit{multi-source} news corpus contains diverse expressions of the same relations, which is a superior to single-source news corpus. Second, a large number of relation mentions can be obtained with a few relation instance seeds. Both aspects benefit training a powerful and robust DS model.
The experimental results show the superiority of our proposed \textsc{Time-DS} for relation extraction. It is worthwhile to highlight our contributions as follows:

\begin{itemize}
\item To alleviate the noises issue of distant supervision in the time-sensitive domain like news data with the timestamp, we take \textit{\textbf{time}} into consideration. 
\item To use \textit{\textbf{time}} in a sophisticated and flexible way, we use curriculum learning in terms of a time series \textit{instance-popularity}, which is proved to be effective for noises elimination.
\item A \textit{multi-source} news corpus with timestamp is collected. Such \textit{multi-source} corpus is more natural to utilize rich temporal statistics compared with independent single documents only.

\end{itemize}

\section{Problem Statement}
In this section, we firstly introduce some concepts used in this paper, then formally define \textsc{Time-DS} for relation extraction. 

\noindent\textsc{\textbf{Definition 1.}} \textbf{(Relation Instance)} \textit{If a relation $r$ holds between two entities $(e_i, e_j)$, we take $r(e_i, e_j)$ as a relation instance, such as \textsc{Partnership}(Microsoft, Facebook).}

\noindent\textsc{\textbf{Definition 2.}} \textbf{(Relation Mention)}
\textit{For the relation instance $r(e_i, e_j)$, we define relation mention as a triple $(e_i, e_j, s)$ , consisting of an entity pair $(e_i, e_j)$ and a sentence $s$. The sentence $s$ contains these two entities $e_i$ and $e_j$, and expresses the relation $r$ of the two entities.}

\noindent\textsc{\textbf{Definition 3.}} \textbf{(Supervision Knowledge)} \textit{In the context of distant supervision, supervision knowledge is some supervision signal for automatic dataset annotation to liberate manpower. Relation instances in knowledge bases are usually taken as such supervision knowledge. However, knowledge bases may not be available in some domains for DS. Alternatively, we propose to extract supervision knowledge from news data via high-quality rules in this paper.}


\noindent\textsc{\textbf{Problem.}} \textbf{(\textsc{Time-DS} for Relation Extraction) } \textit{Given an unlabeled corpus with time stamp like news data, \textsc{Time-DS} is supposed to make the most of \textit{\textbf{time}} to produce high-quality annotated dataset automatically, or to eliminate the bad effects of noises produced in the basic DS for relation extraction modeling training. }

Here are three questions which \textsc{Time-DS} must deal with,

\noindent\textsc{\textbf{Question 1.}} \textit{How to obtain supervision knowledge when knowledge base is unavailable in the target domain?}

\noindent\textsc{\textbf{Question 2.}} \textit{How to eliminate the bad effects of noises in obtained relation mentions through alignment of DS?}

\noindent\textsc{\textbf{Question 3.}} \textit{Can we make use of these noises in a reasonable way instead of discarding them simplistically?}

\section{\textsc{Time-DS}}
\label{method}
\begin{figure*}[htb]
	\centering 
	\includegraphics[width=15cm]{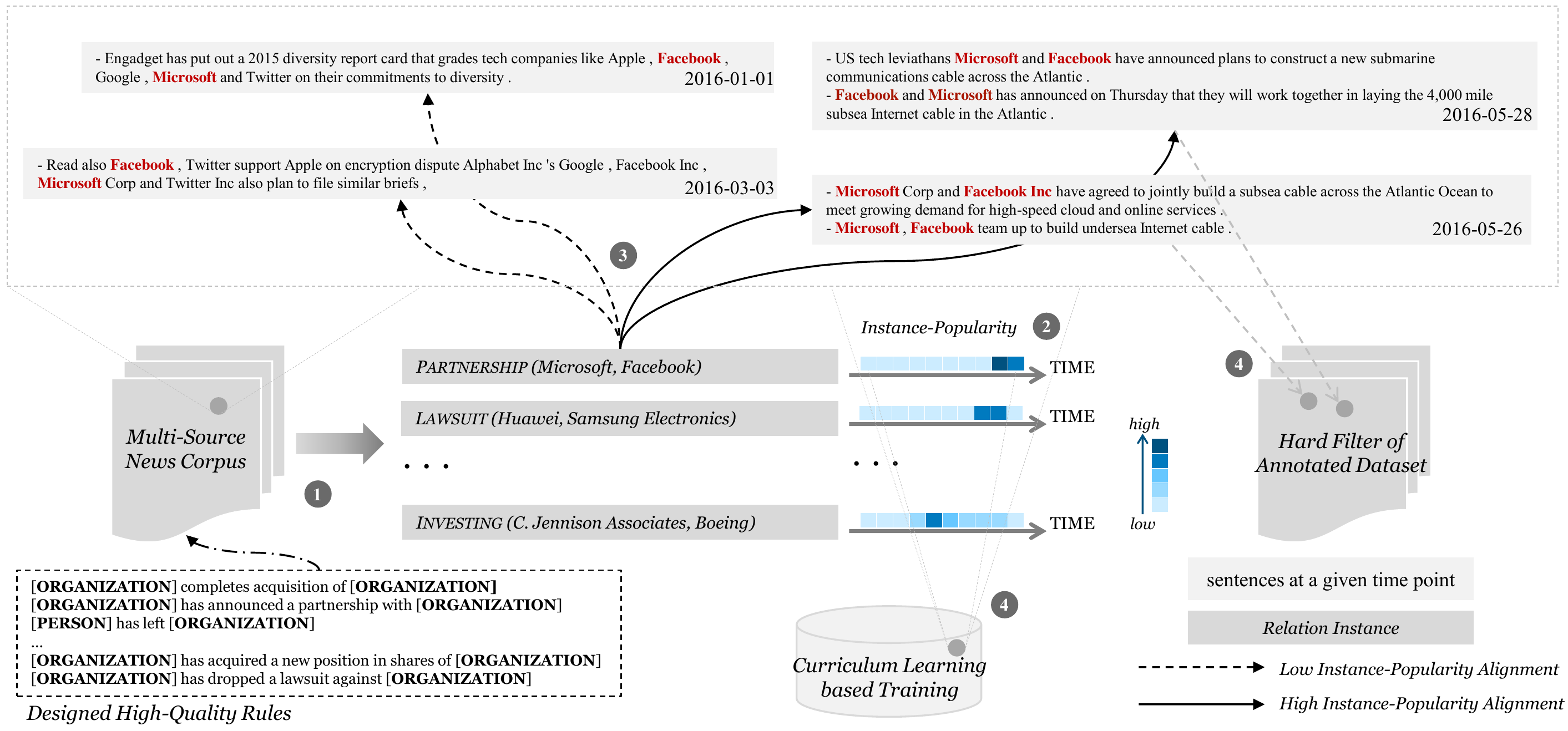}
	\caption{Time-aware distant supervision framework for relation extraction. 
		\textcircled{\scriptsize 1} Extracting supervision knowledge based on high-quality rules.
		\textcircled{\scriptsize 2} Computing of the \textit{instance-popularity} distribution for each relation instance of supervision knowledge.
		\textcircled{\scriptsize 3} Aligning the relation instances to the sentences attached with \textit{instance-popularity}.
		\textcircled{\scriptsize 4} Two strategies, taking \textit{instance-popularity} as a hard filter; Curriculum learning on the weighted training instances, i.e., instances with \textit{instance-popularity}.}
	\label{fig: tade}
\end{figure*}

We introduce details of time-aware distant supervision (\textsc{Time-DS}) framework in this section, following four steps of Figure~\ref{fig: tade}. First, a rule-based method is utilized to extract supervision knowledge for the case when KB is unavailable (in Section~\ref{sko}, to answer \textsc{\textbf{Question 1}}). Second, the \textit{instance-popularity} distribution for each relation instance is approximated based on its rule-matched mentions in news data(in Section~\ref{inspo},~\ref{sec: approx}). Then, aligning the supervision knowledge (i.e., relation instances) to sentences in raw corpus with \textit{instance-popularity} attached, to generate automatic annotated dataset for relation extraction model training (in Section~\ref{strg}). Finally, \textsc{Time-DS} considers two strategies for better use of the time series \textit{instance-popularity}, that is hard filter to answer \textsc{\textbf{Question 2}} and curriculum learning to answer \textsc{\textbf{Question 3}} (in Section~\ref{strg}).

\subsection{Extracting Supervision Knowledge} \label{sko}
KB is a typical supervision knowledge in previous DS based studies. However, KB is usually unavailable in some specific domains. This issue brings the \textsc{\textbf{Question 3}}. Even when KB is available, it is still impossible to get \textit{instance-popularity} distribution only using the relation instances in KB. Some information such as relation mentions and timestamp are also needed. Therefore, it is valuable to get a few of relation instances firstly from raw corpus as supervision knowledge, along with the relation mentions with timestamp.

First, we apply a few of pre-designed high-quality rules to extract relation instances as candidate supervision knowledge. At the same time we reserve the relation mentions with timestamp. Here each rule follows the template of $<$\textbf{Pattern, Constraint}$>$, where Pattern is a regular expression containing a selected connector, Constraint is a lexical constraint on entities to which the pattern can be applied. For example, given the connector ``has formed a partnership with'', we use the pattern ``[entity1] has formed a partnership with [entity2]'' to extract \textsc{Partnership} relationship between organizations, with a constraint that both [entity1] and [entity2] must be organizations. As a consequence, this pattern can match the sentence ``\textit{Microsoft has formed a partnership with Facebook}'', but will not match the sentence ``\textit{Kevin has formed a partnership with Jack to finish the project}''.

Second, we calculate the confidence of each extracted relation instance, and set a reasonable threshold as filter to obtain the final supervision knowledge (i.e. a set of relation instances with high-confidence), as the first step showed in Figure~\ref{fig: tade}. We believe that the confidence of a relation instance $r(e_i, e_j)$, denoted as $C(r(e_i, e_j))$, depends on the amount of relation mentions and the kinds of matched rules in these mentions. We assume that more mentions and matched rules mean more reliable relation instance. According to the assumption and treat mentions and matched rules equally, we define the $C(r(e_i, e_j))$ as follows,
\begin{equation}\label{confidence_def}
\small
C(r(e_i, e_j))= \frac{|\{rule|\, \mathrm{matches}\, r(e_i, e_j)\}|}{Z_{rule}}
+\frac{|\{m|\, \mathrm{mention\_of}\, r(e_i, e_j)\}|}{Z_{m}}
\end{equation}
\noindent where $\{rule|\, \mathrm{match}\, r(e_i, e_j)\}$ represents the set of matched rules for $r(e_i, e_j)$, and $\{m|\, \mathrm{mention\_of}\, r(e_i, e_j)\}$ represents the set of matched sentences for $r(e_i, e_j)$. $Z_{rule}$ and $Z_{s}$ represent the maximum  number of the matched patterns and sentences in all relation instances.

\subsection{Definition of Instance-Popularity}\label{inspo}
The assumption of distant supervision is: any sentence that contains a pair of entities which participate in a relation instance is likely to express that relation in some way. However, this assumption is not always true. A large part of sentences containing a pair of entities are noises. The previous Figure~\ref{fig: insight} indicates that the mentions in a certain period of time have extremely less noises, while in other days are more likely false positives. In other word, whether an aligned sentence expresses the corresponding relation has a strong relevance with \textbf{\textit{time}}. 

Following the intuition, we introduce a time series \textit{instance-popularity} for each relation instance to indicate how many news expressing the relation every certain period of time. Given a sentence at some time point, which contains two entities of a relation instance, we assume that the certainty of expressing the relation is proportional to the \textit{instance-popularity} at that time point of the relation instance. \textit{Instance-popularity} is to prepare for figuring out the issue brought by \noindent\textsc{\textbf{Question 2}}.

Formally, the \textit{instance-popularity} (denoted as InsPo) of a given relation instance $r(e_i, e_j)$ at time $t$ is defined as:
\begin{equation}\label{IP_def}
\small
\textrm{InsPo}_{r(e_i, e_j)}^t=\frac{|A_{r(e_i, e_j)}^t|}{|\Omega_{r(e_i, e_j)}|}
\end{equation}
\noindent where $\textrm{InsPo}_{r(e_i, e_j)}^t$, $A_{r(e_i, e_j)}^t$ represent the \textit{instance-popularity} of $r(e_i, e_j)$ at $t$ and the set of sentences expressing $r(e_i, e_j)$ at $t$-centric time-window separately. $|A_{r(e_i, e_j)}^t|$ denotes the amount of the $A_{r(e_i, e_j)}^t$. $\Omega_{r(e_i, e_j)}$ is the whole set of sentences expressing $r(e_i, e_j)$ over time for normalization, and
\begin{equation}\label{Omega_def}
\small
|{\Omega}_{r(e_i, e_j)}| = \frac{1}{L}\sum_{n=1}^{N}|{A}_{r(e_i, e_j)}^{t_n}|
\end{equation}
\noindent where $|\Omega_{r(e_i, e_j)}|$ is the amount of $\Omega_{r(e_i, e_j)}$. $L$ is the length of the time-window, and $N$ is the amount of time points we concern.

\subsection{Approximation of Instance-Popularity}
\label{sec: approx}
The actual whole set of sentences which express the relation instance is usually unavailable in practice. Thus we cannot compute the \textit{instance-popularity} directly according to Equation~\ref{IP_def}. In this section, we provide an approximate method to calculate the \textit{instance-popularity}.

For each relation instance $r(e_i, e_j)$ in the supervision knowledge (gained in Sec~\ref{inspo}), the set of its whole rule-matched sentences is denoted as ${\Omega'}_{r(e_i, e_j)}$, which is a sub-set of $|\Omega_{r(e_i, e_j)}|$. Further, we can obtain the sentences set in any $t$-centric time-window from ${\Omega'}_{r(e_i, e_j)}$, denoted as ${A'}_{r(e_i, e_j)}^t$. The assumption is that people select relation patterns under some distribution in news data, and we use $p_{\mathcal{P}}^{t}$ to denotes such probability of the relation pattern $\mathcal{P}$ being selected to express the given relation instance at time point $t$. Then we can calculate $|{A'}_{r(e_i, e_j)}^t|$ and $|{\Omega'}_{r(e_i, e_j)}|$ under the probability distribution of the relation patterns in the pre-designed rules, 

\begin{equation}\label{A}
\small
|{A'}_{r(e_i, e_j)}^t| = \sum_{k=1}^{K}|{A}_{r(e_i, e_j)}^{t}|\cdot p_{\mathcal{P}_k}^t
\end{equation}
\begin{equation}\label{B}
\small
|{\Omega'}_{r(e_i, e_j)}| = \frac{1}{L}\sum_{n=1}^{N}|{A'}_{r(e_i, e_j)}^{t_n}|
\end{equation}

\noindent where $\mathcal{P}_k$ is the $k$-th relation pattern. $K, N$ are the number of relation patterns and time points separately. 


\textbf{\textsc{Assumption.}} \textit{In the \textit{multi-source} news corpus, a given a pattern $\mathcal{P}_k$ for expressing a relation instance $r(e_i, e_j)$ would be selected with the same probability $p_{\mathcal{P}_k}$ at any time, which means that the probability of a given pattern being selected is independent of time. Therefore we have, $p_{\mathcal{P}_k}^{t_1} \approx p_{\mathcal{P}_k}^{t_2} \approx \cdots \approx p_{\mathcal{P}_k}$.}

According to the Equation~\ref{IP_def} to ~\ref{B}, and \textbf{\textsc{Assumption}}, we can get the following equations.
\begin{equation}\label{AA}
\small
|{A}_{r(e_i, e_j)}^t| \approx \frac{1}{\sum_{k=1}^{K}p_{\mathcal{P}_k}}\cdot |{A'}_{r(e_i, e_j)}^t|
\end{equation}
\begin{equation}\label{BB}
\small
|{\Omega}_{r(e_i, e_j)}| \approx \frac{1}{\sum_{k=1}^{K}p_{\mathcal{P}_k}}\cdot |{\Omega'}_{r(e_i, e_j)}|
\end{equation}

Based on the Equation~\ref{AA},~\ref{BB}, we can approximate the \textit{instance-popularity} for $r(e_i, e_j)$ at time point $t$, as the second step in Figure~\ref{fig: tade}. The approximation equation is as follows. 
\begin{equation}\label{IP_cal}
\small
\textrm{InsPo}_{r(e_i, e_j)}^t \approx \frac{|{A'}_{r(e_i, e_j)}^t|}{|{\Omega'}_{r(e_i, e_j)}|}
\end{equation}

\subsection{Two Strategies to Exploit Instance-Popularity} \label{strg}
Given the unlabeled corpus with timestamp, we align the supervision knowledge, i.e., relation instances, to the corpus, as the third step showed in Figure~\ref{fig: tade}. A lot of relation mentions can be obtained, along with the corresponding approximate \textit{instance-popularity}. In other words, a large scaled annotated dataset attached with \textit{instance-popularity} is acquired. Relation extraction model can be trained on the datasets. In the training process of relation extraction model, for better use of the time series \textit{instance-popularity}, \textsc{Time-DS} considers two strategies, as the last step of Figure~\ref{fig: tade}, that is hard filter and curriculum learning. 

\textbf{Hard Filter.} We have obtained the dataset attached with \textit{instance-popularity} in sentence-level, and \textit{instance-popularity} is to quantize the reliability of the annotated sentences. Hard filter sets a hard threshold of the \textit{instance-popularity} to filter the dataset to get a higher-quality sub-dataset, discarding the noises. This is a simplest way to utilize \textsc{Time-DS}

However, despite the effectiveness of hard filter, It (1) heavily relies on the threshold of \textit{instance-popularity}, and (2) refuses to use some noises to make DS based models robust. We hope that the noises can be used reasonably to get a more sufficient training rather than discard these noises directly, as asked in \noindent\textsc{\textbf{Question 3}}. A natural idea is guiding the relation extraction model to adapt to the noisy training sets gradually, i.e., learning something simple first, and then attempting to deal with noises. Fortunately, a technique called \textit{curriculum learning} fits our problem.

\textbf{Curriculum Learning.} The main idea of curriculum learning~\cite{bengio2009curriculum} is simple: starting with the easiest aspect of a task, and leveling up the difficulty gradually. In our study, we begin with the high-quality annotated sentences and gradually add low-quality sentences into training set according to \textit{instance-popularity}. In particular, all the annotated sentences from distant supervision are ranked by \textit{instance-popularity} from high to low. Then we divide the ranking list into several groups by assigning different thresholds of \textit{instance-popularity}, i.e., \{Rank$_{1,...,n}$, Rank$_{n+1,...,n+k}$, Rank$_{n+k+1,...,n+m}$, ...\}. Therefore, different training sets can be easily created by gradually combining different groups of annotated sentences with the ranking order, i.e.,  Rank$_{1,...,n}$, Rank$_{1,...,n+k}$, Rank$_{1,...,n+m}$, etc.

Then, following the strategy of curriculum learning, (1) first, the model is trained on the highest-quality training set, that is Rank$_{1,...,n}$. After the training is complete, (2) the second highest-quality training set and the previous training set, i.e, the highest-quality training set, are merged to generate a new training set, that is Rank$_{1,...,n+k}$. The model is trained again in this new training set. (3) Then, repeat the above processes, i.e., add the the lower-quality training set gradually and train the model in every new training set until all annotated sentences from DS are taken into consideration. Note that the training instances, i.e., sentences, are shuffled during each training process.

\section{Experiments}

In this part, we conduct experiments of relations extraction/classification on five time-sensitive relations from a \textit{multi-source} news corpus. 

\begin{table*}[htb]
	\centering
    \small
	\renewcommand{\arraystretch}{1.2}
	\setlength\tabcolsep{3pt}
	\begin{tabular}{|c|c|c|c|c|c|c|c|c|c|c|}
		\hline
		\multicolumn{1}{|c|}{\multirow{2}[1]{*}{\textbf{Relation}}} & \multicolumn{2}{|c|}{\textbf{\#Relation Instances}} &
		\multicolumn{7}{|c|}{\textbf{Training Set} \textbf{\#Sentences}} &
		\multirow{2}{*}{\textbf{Test Set \#Sentence}}\\
		\cline{2-10}
		&\textbf{for Training}&\textbf{for Test}& {Original Set ($\geq 0.0$)} & {$\geq  0.1$} & {$\geq 0.2$} & {$\geq 0.3$} & {$\geq 0.4$} & {$\geq 0.5$} & {$\geq 0.6$} & \multirow{2}{*}{}\\
		\hline
		\textsc{Acquisition} &30&8& 39,365 & 9,905 & 8,361 & 6,963 & 6,388 & 5,825 & 4,664 & 694 \\
		\textsc{Investing} &46&11& 2,741 & 1,227 & 239  & 146  & 142  & 141  & 138  & 48 \\
		\textsc{JobChange} &71&12& 188,945 & 35,154 & 25,177 & 21,622 & 18,655 & 16,452 & 13,507 & 905 \\
        \textsc{Lawsuit} &15&3& 16,503 & 2,147 & 1,588 & 1,334 & 944  & 854  & 697  & 313 \\
		\textsc{Partnership} &12&5& 1,408 & 794  & 794  & 794  & 794  & 794  & 259  & 503 \\ \hdashline
		\textit{total} &174&39& 248,872 & 49,224 & 36,156 & 30,859 & 26,923 & 24,066 & 19,265 & 2,463/376 \\
		\hline
	\end{tabular}%
	\caption{Statistics of the supervision knowledge and annotated dataset. The statistics of the supervision knowledge is reported in the ``\#Relation Instances'' column. ``Original Set'' is the whole original training set, i.e., the training set produced by DS. The other filtered training sets (``$\geq 0.1$'' to ``$\geq 0.6$'') are produced by hard filter, based on different threshold of \textit{instance-popularity}.}
    \vspace{-0.1in}
	\label{tab:annotated dataset}
\end{table*}


\subsection{Data Preparation} \label{corpus}
We collect about 42 million news articles from 50,428 different on-line news websites, and the time spans 8 months from Jan. 2016 to Aug. 2016. In each article, the title, first paragraph, and timestamp are remained to construct a \textit{multi-source} news corpus. The corpus contains nearly 320 million sentences in total\footnote{\scriptsize The multi-source news corpus will be open available.}. Stanford CoreNLP tool~\cite{manning2014stanford} is applied to recognize named entities in the multi-source new corpus. Organization management is an interesting and informative domain in news data, thus we focus on five typical time-sensitive relations in the domain, namely, \textsc{Acquisition}, \textsc{Investing}, \textsc{JobChange}, \textsc{Lawsuit}, and \textsc{Partnership}. It is worth to mention that \textsc{Time-DS} can be easily transfered to any other time-sensitive relations such as \textsc{MarriedTo}, \textsc{VisitIn}, by just designing a few of high-quality rules of these relations.

\textsc{\textbf{Acquisition.}} An organization buys another organization (directed). 
Example: \textit{Verizon} announced it had completed the \$ 4.4 billion acquisition of \textit{AOL}.

\textsc{\textbf{Investing.}} An organization puts money into another organization (directed). 
Example: \textit{Vontobel Asset Management Inc.} boosted its position in shares of \textit{Mastercard Inc.}.

\textsc{\textbf{JobChange.}} A person leave or join in an organization (directed). 
Example: \textit{Papiss Cisse} has left \textit{Newcastle United}.

\textsc{\textbf{Lawsuit.}} An organization suits another organization (directed). 
Example: \textit{Samsung Elec} sues \textit{Huawei} for patent infringement.

\textsc{\textbf{Partnership.}} An organization forms a partnership with another organization (undirected). 
Example: \textit{Konami} has announced a partnership with \textit{FC Barcelona} for PES 2017.

\textbf{Test Set.} We follow the previous study~\cite{mintz2009distant} to hold out part of the relation instances in supervision knowledge to be aligned into corpus to get test set. However, such test set also suffers from the wrong labeling problem, leading to a rough measure of the performance. Hence we refine the test set in three steps. (1) First, filter the test instances with a suitable \textit{instance-popularity} threshold\footnote{\scriptsize In our experiments, we set the threshold as 0.2 for relation \textsc{Investing} and 0.7 for the other four relations} to get the candidate positive samples, and also reserve some of filtered instances as candidate negative samples. (2) Then three experts in the relevant domain proofread the candidate test set independently, that is judging and correcting the correctness of the existing tags. (3) Finally, the remaining disagreements are resolved, and if no consensus could be achieved, the samples are removed. At last, 2,463 positive and 376 negative are achieved to form the final test set (in Table~\ref{tab:annotated dataset}), in which 186 samples has been corrected. 

\textbf{Validation Set.} The above test set is randomly partitioned into 10 equal size subsets. Of the 10 subsets, a single subset is retained as the validation set for selecting model, and the remaining 9 subsets are used as testing data. This process is then repeated 10 times, with each of the 10 subsets used exactly once as the validation set. The trained model with the best averaged performance on validation sets is selected as the final model for evaluation.


\subsection{Target Models}

In this part, we describe two models which are fed into our \textsc{Time-DS} and the basic DS framework for end-to-end relation extraction task and relation classification task. The relation extraction task is to extract relation mentions from the given sentences, and categorize them into a pre-defined set 
of relation types. If the relation mentions are given, then the task is a classification problem, 
called relation classification. In this paper, the relation extraction and classification we studied 
are both in sentence-level.

\textbf{End-to-End Relation Extraction.} We feed the model proposed by Zheng et al~\shortcite{zheng2017joint}, i.e., LSTM-LSTM-Bias into our \textsc{Time-DS} framework. LSTM-LSTM-Bias designs a novel tagging schema to convert the task to a sequence tagging problem. Therefore the model can extract entities and relations jointly without other redundant information and achieve the best results on the public dataset. Since LSTM-LSTM-Bias is a sequence tagging model, the training only need word-level positive and negative. Thus we can train the model on the automatic annotated datasets generated by \textsc{Time-DS} directly.

\textbf{Relation Classification.} We applied a Bi-LSTM and Attention based neural model proposed by Zhou~\shortcite{zhou2016attention}, which is a typical paradigm for relation classification, called Att-BLSTM. The training of Att-BLSTM needs negative samples, which is unavailable from the generated datasets in manner of \textsc{Time-DS}. To obtain negative samples, we replace tail entity of each relation instance with another entity, which is in the same sentence. For instance, given a sentence ``\textit{A has formed a partnership with B, which is located in C}'' and its relation instance \textsc{Partnership}(\textit{A, B}), we replace the entity B with entity C. The new relation instance \textsc{Partnership}(\textit{A, C}) and original sentence form a negative sample. 

\textbf{Metrics.} Similar with previous work, we report the aggregate precision/recall curves on the end-to-end relation extraction model, and macro-F1 on relation classification model.

\subsection{Annotated Datasets Generation with Instance-Popularity}\label{sec: distant supervision}
For \textsc{\textbf{Question 1}}, i.e., how to obtain supervision knowledge when knowledge base is unavailable, we apply only a few of manually designed high-quality rules (see in Section~\ref{sko})\footnote{\scriptsize Only 5 to 8 different rules for each relation is suitable in our experiments.}. Meanwhile the rule-matched sentences are reserved for \textit{instance-popularity} approximation. Then we use confidence (see Equation~\ref{confidence_def}) to filter the rule-based extracted relation instances to form the final supervision knowledge. The final supervision knowledge is divided into two sub-sets, i.e., for training and test, according to held-out method. The statistics is reported in the ``\#Relation Instances'' column of Table~\ref{tab:annotated dataset}. We can see that the amount of gained relation instances is much smaller (only about 2.1 hundred) compared with the existent KB(over 3.2 million relation instances are obtained in FreeBase in~\cite{riedel2010modeling}). However, with the help of the \textit{multi-source} news corpus, we can gain huge amount of expressive annotated sentences.

Aligning the relation instances to the corpus, we get 248,872 training instances, i.e., relation mentions, annotated with \textit{instance-popularity}, called \textit{Original Set}. The \textit{Instance-popularity} distribution of relation instance is directly approximated according to Equation~\ref{IP_cal}\footnote{\scriptsize The size of time-window is set 3 days in our experiments}, based on the reserved rule-matched sentences. To use hard filter strategy, we filter the \textit{Original Set} to get another six sub-sets according to the different \textit{instance-popularity} thresholds. The highest threshold is set 0.6, because the amount of \textsc{Investing} mentions are nearly zero when the threshold higher than 0.6. The overall statistic of annotated dataset is showed in Table~\ref{tab:annotated dataset}. 

\subsection{Effectiveness of Hard Filter}

To answer \textsc{\textbf{Question 2}}, i.e., how to eliminate the bad effects of noises produced in the basic DS, we adapt a \textit{instance-popularity}-based hard filter strategy. Here we examine the the performance of hard filter strategy on relation extraction and classification tasks. 

\begin{figure}[htb]
	\centering 
	\setlength{\abovecaptionskip}{0pt}
	\includegraphics[width=8.5cm]{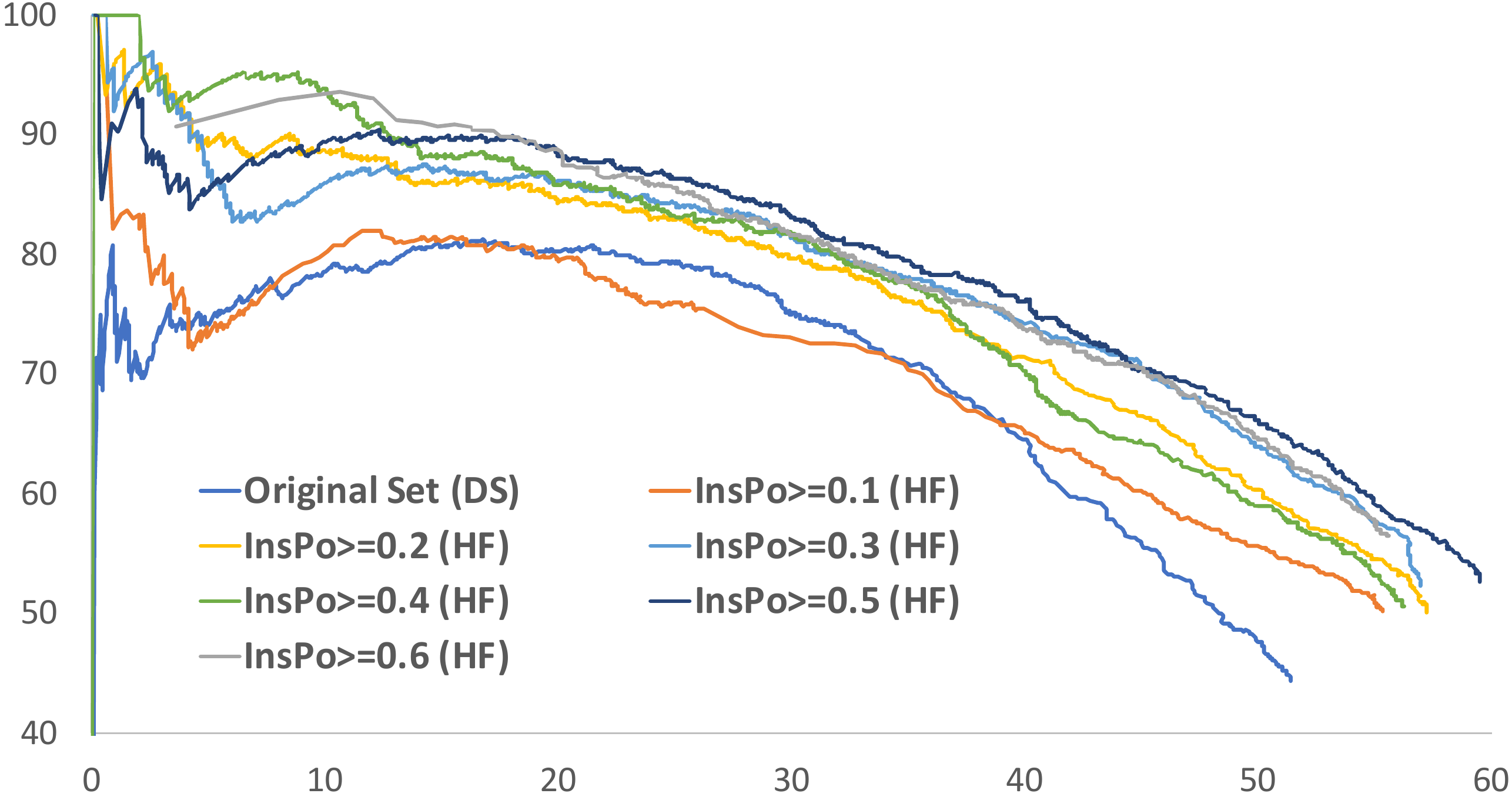}
	\caption{Aggregate precision/recall curves of relation extraction model trained on different datasets. The \textit{Original Set} curve represents the performance of training on the annotated dataset generated by the basic DS. The other six curves represent the performance of training on the annotated datasets generated by different hard filters, denoted as ``HF''.}
	\label{fig: re_exp1}
      \vspace{-0.1in}
\end{figure}

Figure ~\ref{fig: re_exp1} shows the aggregate precision/recall curves of relation extraction model trained on different datasets. we find that: (1) Except 0.1 threshold of hard filter, the model trained on all datasets of hard filter outperform the one trained in manner of the basic DS, i.e., Original Set. (2) It is clear that the accuracy/recall performance increases with the increase of the threshold of \textit{instance-popularity} when the threshold is lower than 0.5. 

\begin{figure*}[htb]
	\centering 
	\setlength{\abovecaptionskip}{0pt}
	\includegraphics[width=16cm]{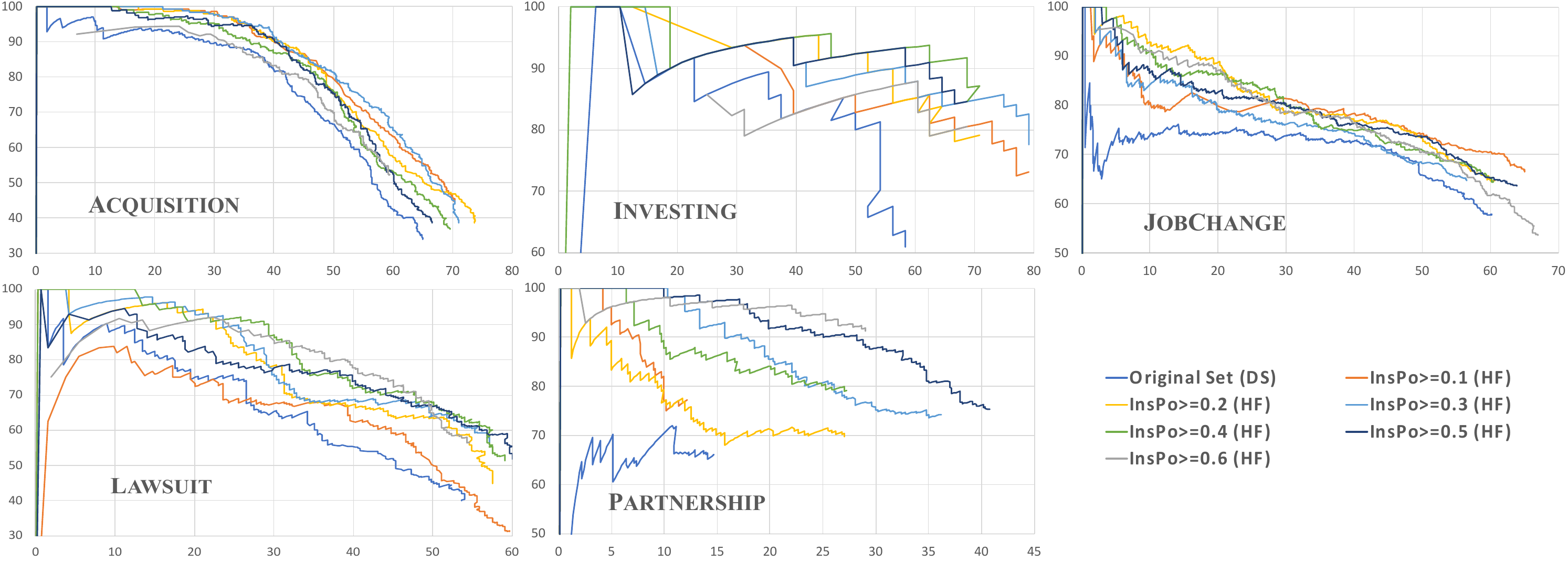}
	\caption{More fine-grained precision/recall curves on the five target relations respectively.}
	\label{fig: re_exp2}
\end{figure*}

We also investigate more fine-grained precision/recall curves of the five relations separately (see in Figure~\ref{fig: re_exp2})\footnote{\scriptsize Some precision/recall curves backtrack, such as \textsc{Investing}, that is because LSTM-LSTM-Bias is trained and to do predict in tag-level instead of relation-level~\cite{zheng2017joint}.}. we find that: (1) Similar general tendency is observed, that is the model train on datasets of hard filter outperform the one trained in manner of the basic DS when threshold of hard filter higher than some value. (2) The effects of hard filters on different relation types is not exactly same. The performance for \textsc{Partnership} increases as the hard filter became stricter, while the performance on \textsc{JobChange} already peaks when the threshold arrives 0.1. 

Table~\ref{tab: rec_exp1} presents the macro-average F1 value on relation classification model. We can see that (1) \textsc{Time-DS} framework on different fractions of training set outperforms the basic DS framework on the whole training set for relation classification; (2) The macro-F1 value tends to increase with the increase of the threshold of \textit{instance-popularity}; (3) Even with a smaller scale of training data, when $InsPo\geq 0.6$ the corresponding training set is almost one-thirteenth of the \textit{Original Set} (see Table~\ref{tab:annotated dataset}), the \textsc{Time-DS} framework outperforms basic DS framework, achieving significant improvement for relation classification. Above all, it is very clear from these three observations that hard filter with \textit{instance-popularity} is a very effective strategy to segment \textit{Original Set} and eliminate the bad effects of the noises generated from the basic DS.



\begin{table}[htb]
	\begin{center}
    	\small
		\renewcommand{\arraystretch}{1.4}
		\setlength\tabcolsep{3pt}
		\begin{tabular}{|l|r|r|r|}
			\hline
			\bf Training Set & \bf macro-P(\%) & \bf macro-R(\%) & \bf macro-F1(\%) \\ \hline
            \multicolumn{4}{|l|}{the Basic DS} \\ \hdashline
			Original Set & $67.62 \pm 0.27$ & $\mathbf{78.62 \pm 0.26}$ &	$72.71 \pm 0.25$ \\ \hline
            \multicolumn{4}{|l|}{\textsc{Time-DS} with Hard Filter} \\ \hdashline
			InsPo$\geq 0.1$ & $70.72 \pm	0.39$ & $77.04 \pm	0.20$ & $73.74 \pm	0.27$ \\
			InsPo$\geq 0.2$ & $72.73 \pm	0.33$ & $75.96 \pm	0.28$ & $74.31 \pm	0.27 $ \\
			InsPo$\geq 0.3$ & $\mathbf{76.38 \pm	0.23}$ & $77.65 \pm	0.27$ &$\mathbf{77.01 \pm	0.20} $ \\
			InsPo$\geq 0.4$ & $71.74 \pm	0.29$ & $74.30 \pm	0.25$ & $73.00 \pm	0.22 $ \\
			InsPo$\geq 0.5$ & $72.35 \pm	0.39$ & $75.27 \pm	0.33$ & $73.78 \pm	0.31 $ \\
			InsPo$\geq 0.6$ & $74.27 \pm	0.37 $ & $77.76 \pm	0.32$ & $75.97 \pm	0.31 $ \\
			\hline
		\end{tabular}
	\end{center}
	\caption{\label{tab: rec_exp1} Macro-average of precision, recall and F1 of relation classification model, which is trained on the \textit{Original Set} and the other six filtered training sets respectively.}
   \vspace{-0.4in}
\end{table}

\begin{figure*}[htb]
	\centering 
	\setlength{\abovecaptionskip}{0pt}
	\includegraphics[width=17.5cm]{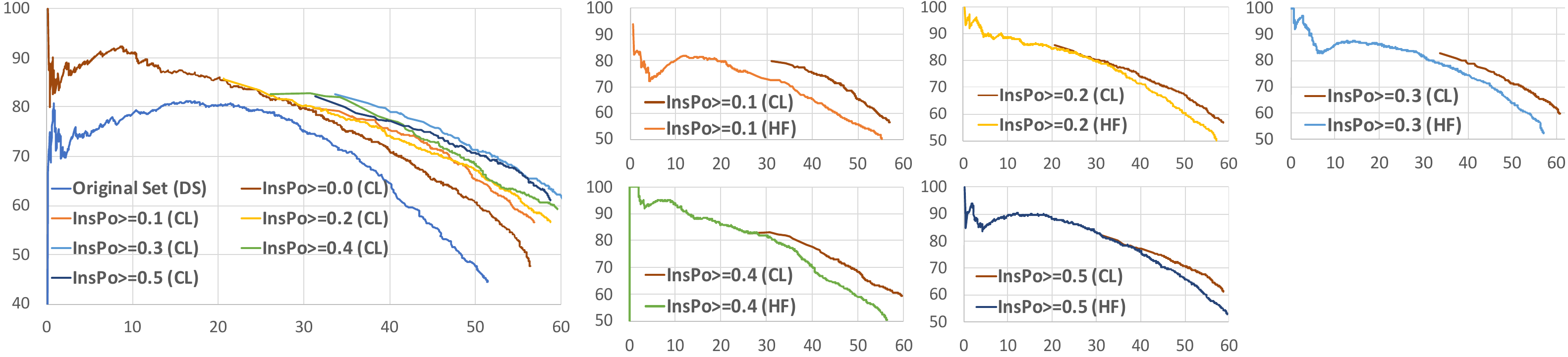}
	\caption{Aggregate precision/recall curves of relation extraction model trained on six training sets with traditional training in hard filter and curriculum learning, where ``CL'' represents curriculum learning.}
	\label{fig: re_exp3}
\end{figure*}

\begin{table}[htb]
	\begin{center}
    	\small
		\renewcommand{\arraystretch}{1.4}
		\setlength\tabcolsep{3pt}
		\begin{tabular}{|l|r|r|r|}
			\hline
			\bf Training Set & \bf macro-P(\%) & \bf macro-R(\%) & \bf macro-F1(\%) \\ \hline
            \multicolumn{4}{|l|}{the Basic DS} \\ \hdashline
			Original Set & $67.62 \pm 0.27$ & $ 78.62 \pm 0.26$ &	$72.71 \pm 0.25$ \\ \hline
            \multicolumn{4}{|l|}{\textsc{Time-DS} with Curriculum Learning} \\ \hdashline
			InsPo$\geq 0.0$ (7th round) & $69.09\pm 0.39$ &	$77.93 \pm 0.24$ & $73.24 \pm 0.29 $ \\ 
			InsPo$\geq 0.1$ (6th round) & $72.41 \pm 0.33 $ & $\mathbf{79.76\pm 0.42} $ & $75.91\pm 0.31 $ \\
			InsPo$\geq 0.2$ (5th round) & $75.75\pm 0.44$ & $76.50\pm 	0.24$ & $76.12 \pm 	0.28 $ \\
			InsPo$\geq 0.3$ (4th round) & $\mathbf{79.04\pm 0.29}$ & $77.73\pm 	0.24$ & $\mathbf{78.38 \pm 	0.20}$ \\
			InsPo$\geq 0.4$ (3rd round) & $76.53\pm 0.25$ & $75.52\pm 	0.26$ & $76.02 \pm 	0.19 $ \\
			InsPo$\geq 0.5$ (2nd round) & $78.34 \pm 0.28$ & $76.06 \pm	0.29$ & $77.18 \pm	0.21 $ \\
            InsPo$\geq 0.6$ (1st round) & $74.27 \pm	0.37 $ & $77.76 \pm	0.32$ & $75.97 \pm	0.31 $ \\
			\hline
		\end{tabular}
	\end{center}
    \caption{\label{tab: rec_exp2} Macro-average of precision, recall and F1 of relation classification model trained on \textit{Original Set} with curriculum learning.}
    \vspace{-0.4in}
\end{table}

\subsection{Effectiveness of Curriculum Learning}

Hard filter strategy heavily relies on the threshold settings to remove the effects of noisy samples. However, noisy samples still play important roles in improving the robustness of models. This issue naturally bring the \textsc{\textbf{Question 3}}, i.e., can we make use of these noises in a reasonable way to improve the robustness rather than discarding them simplistically. The strategy of curriculum learning is used in \textsc{Time-DS} to answer \textsc{\textbf{Question 3}}.

To implement \textsc{Time-DS} with curriculum learning, we reconstruct the original set to get 7 subsets and apply a curriculum learning strategy. In particular, we distribute the original set to get seven subsets with different \textit{instance-popularity} ranges, i.e., [0.6, 1.0], [0.5, 1.0], [0.4, 1.0], [0.3, 1.0], [0.2, 1.0], [0.1, 1.0], [0.0, 1.0]. At 1st round, we use the subset [0.6, 1.0] to train the model for relation classification. At 2nd round, we use the subset [0.5, 1.0] to retrain the model on the basis of the model obtained from the 1st round. At 3rd round, we use the subset [0.4, 1.0] to retrain the model on the basis of the model obtained from the 2nd round. The latter rounds follow the same rules until 7th round when all instances of \textit{Original Set} participate in the training process.

Figure~\ref{fig: re_exp3} presents the performance on relation extraction\footnote{\scriptsize Some precision/recall curves starts from non-zero point, which is because \textit{softmax} may output 1 in some dimension for optimization in Python language.}. We find that (1) curriculum learning based \textsc{Time-DS} significantly outperform the basic DS in any training round; (2) Each training round of curriculum learning outperform the traditional training in hard filter strategy. Table~\ref{tab: rec_exp2} presents the performance on relation classification. (1) From the comparison between the different rounds of \textsc{Time-DS} with curriculum learning and the basic DS, it is clear that \textsc{Time-DS} with curriculum learning outperforms the basic DS in every round of training and \textsc{Time-DS} with curriculum learning achieves the best performance at round 4. (2) From round 1 to round 4, the noisy samples are gradually added, the performance tends to increase. However, the performance decrease when adding to much noisy samples into the training set after round 4.


\subsection{Deep analysis of Instance-Popularity}
In this section, we provide deep analysis about how \textit{instance-popularity} works to encode the strong relevance of timestamp and the true relation mention in news data.

\textbf{Time-sensitive Relations.} Many relations in news is very sensitive to time, such as these relations in Table~\ref{tab: InsPo Cases}. At the time period of the establishment of these relations, there would be extremely less noises of these relation mentions. This feature would benefit a lot for the alignment process of distant supervision. 
Therefore, we want to check the consistency of peaking time of \textit{instance-popularity} and the establishment time of each relation instance. In particular, we sample several relation instances randomly, and acquire the establishment time from Wikipedia or news report. In Table~\ref{tab: InsPo Cases}, we can find that the time when \textit{instance-popularity} reaches peak is usually consistent with the establishment time. Therefore, it is reasonable to take \textit{instance-popularity} as measure to find relation mentions as training data with less noises.

\begin{table}[htb]
	\begin{center}
    	\small
		\renewcommand{\arraystretch}{1.4}
		\setlength\tabcolsep{3pt}
		\begin{tabular}{|r|c|c|c|c|c|c|c|}
			\hline
			\bf  & \bf Orig. Set & $\mathbf{\geq 0.1}$ & $\mathbf{\geq 0.2}$ & $\mathbf{\geq 0.3}$ & $\mathbf{\geq 0.4}$& $\mathbf{\geq 0.5}$ & $\mathbf{\geq 0.6}$\\ \hline
			\bf Noise Ratio & 0.66 & 0.37 & 0.33 & 0.23 & 0.37 & 0.21 & 0.18 \\ 
            \bf Set Scale & 248,872 & 49,224 & 36,156 & 30,859 & 26,923 & 24,066 & 19,265 \\ \hline
		\end{tabular}
	\end{center}
    \caption{\label{tab: inspo_study} Noise ratio and set scale of the different training sets.}
    \vspace{-0.1in}
\end{table}

\begin{table*}[ht]
	\centering
	\small
	\renewcommand{\arraystretch}{1.1}
	\setlength\tabcolsep{3pt}
	\begin{tabular}{|l|l|c|c|}
		\hline
		\textbf{Relation Instance} & \textbf{Instance-Popularity} & \textbf{Peaking Time} & \textbf{Establishment Time} \\
		\hline
		\textsc{Acquisition} \textit{(Pfizer, Anacor Pharmaceuticals)}
		&\includegraphics[scale=0.9]{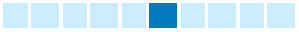} May&May 16-18, 2016 &May 16, 2016\\
		\textsc{Acquisition} \textit{(NBC, DreamWorks Animation)}&\includegraphics[scale=0.9]{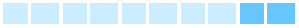} Apr.&Apr. 25-30, 2016 &Apr. 28, 2016\\
		\textsc{Investing} \textit{(Private Trust, Honeywell International)}
		&\includegraphics[scale=0.9]{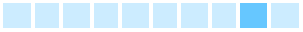} Feb.& Feb. 25-27 2016 & Feb. 25 2016 \\
		\textsc{Investing} \textit{(Jennison Associates, Boeing)}&\includegraphics[scale=0.9]{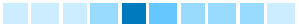} Jan.&Jan. 13-15, 2016 & Jan. 13, 2016\\
		\textsc{JobChange} \textit{(Louis Van Gaal, Manchester United)}
		&\includegraphics[scale=0.9]{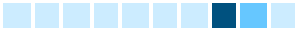} May&May 22-24, 2016 &May 23, 2016\\
		\textsc{JobChange} \textit{(Derek Fisher, Knicks)}&\includegraphics[scale=0.9]{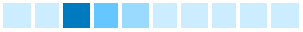} Feb.&Feb. 7-9, 2016 &Feb. 7, 2016\\
		\textsc{Lawsuit}\textit{(Huawei, Samsung Electronics)}
		&\includegraphics[scale=0.9]{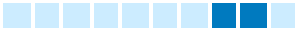} May&May 22-27, 2016 &May 25, 2016\\
		\textsc{Lawsuit} \textit{(Wal-Mart Stores Inc., Visa Inc)}&\includegraphics[scale=0.9]{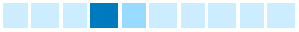} May&May 10-12, 2016 &May 10, 2016\\
		\textsc{Partnership}\textit{ (Microsoft, Facebook)}
		& \includegraphics[scale=0.9]{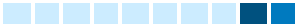} May& May 25-27, 2016 & May 26, 2016\\
		\textsc{Partnership} \textit{(Google, Fiat)}&\includegraphics[scale=0.9]{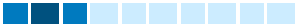} May&May 4-6, 2016 &May 3, 2016\\
		\hline
	\end{tabular}
	\caption{\textit{Instance-popularity} cases of some relation instances, and their establishment time. Due to the limited space, we only display the month when \textit{instance-popularity} peaks and merge three continuous days to be displayed in one cell.}
	\vspace{-0.3in}
	\label{tab: InsPo Cases}
\end{table*}

\textbf{The Ability to Eliminate the Noises. } 
We investigate the ability of \textit{instance-popularity} as hard filter to eliminate the noises generated from the DS alignment. Although the \textit{instance-popularity} has been proved to be useful indirectly a component of \textsc{Time-DS}, a direct evaluation would be much more straightforward. Specifically, we randomly sample 100 training instances respectively from each data sets to check the ratio of noises in each training set. Table~\ref{tab: inspo_study} presents the the ratio of noises in each training set. Note that our models which are trained on subset ``$\geq 0.3$'' and subset ``$\geq 0.5$'' achieve the best performance for relation classification and extraction. It is clear that the training sets with the lower ratio of noises tend to be those with which our model achieves the better performance. This is not very strict because the scale of training set also affects the performance.


\textbf{Error Cases Study.} In this study, we have two types of error cases. First, there are some aligned sentences with low-\textit{instance-popularity} is actually the true relation instances. We call them low-\textit{instance-popularity} but positive case (LP). Second, there are some aligned sentences with timestamp consisting with the peak time of \textit{instance-popularity} are actually the fake relation instances. We call them high-\textit{instance-popularity} but negative case (HN).
We present some LP and HN cases as follows.

\textbf{LP1}. ``\textit{Jose Mourinho} is reportedly set to be confirmed as \textit{Manchester United}'s new manager in the coming days.'', InsPo: 0.0, relation: \textsc{JobChange}.

\textbf{LP2}. ``\textit{Activision Blizzard}'s acquisition of \textit{Major League Gaming} appears to be bearing fruit.'', InsPo: 0.053, relation: \textsc{Acquisition}.

\textbf{LP3}. ``... between two giants in the technology world, as we 've seen repeatedly with the \textit{Apple} v. \textit{Samsung} litigation .'', InsPo: 0.097, relation: \textsc{Lawsuit}.

\textbf{HN1}. ``\textit{LinkedIn} will give \textit{Microsoft} an even greater foothold in the space ...'', InsPo: 0.67, relation: \textsc{Acquisition}.

\textbf{HN2}. ``\textit{Google} and \textit{Fiat} Chrysler engineers will fit Google 's autonomous driving technology into the Pacifica minivan.'', InsPo: 0.5, relation: \textsc{Partnership}.

\textbf{HN3}. ``Warren Buffett, fondly known as the Oracle of Omaha ... behind brainchild \textit{Berkshire Hathaway Inc.} just upped his stake in \textit{Apple Inc.} by a significant chunk.'', InsPo: 0.80, relation: \textsc{Investing}.

The above LP cases have different situations. (1) Some relation instances are reported in the news before the official establishment of the relations. Therefore, the timestamp of these mentions is earlier than the peak time of \textit{instance-popularity}, e.g., LP1. 
(2) Some relation instances are still reported in the news even a long time after the official establishment of the relations. Therefore, the timestamp of these mentions is much latter than the peak time of \textit{instance-popularity}, e.g., LP2. (3) Some relation instances will be reported by news media for a long time. In this case, \textit{instance-popularity} will be normally distributed in a long time period. Therefore, \textit{instance-popularity} usually fails to detect the time period of the establishment of the relations in a short time interval, e.g., the 6 years lawsuits between Apple Inc. and Samsung Electronics.

The above HN cases have different situations. (1) Some alignment mentions actually talk about other aspects of the mentioned entities. For example HN1 discusses the influence of the acquisition rather than expressing the acquisition as a relation. (2) Some cases provide incomplete information, making it hard to confirm the existence of the relation, e.g., HN2 and HN3. 


\section{Related Work}
\subsection{Improvements for Distant Supervision}

To alleviate effects of noises in automatic annotated dataset of DS, some studies captured certain types of noise and aggregated multi-instance learning~\cite{riedel2010modeling,surdeanu2012multi,ritter2013modeling,min2013distant}. Some neural networks methods learned from multiple instances attentively, without explicitly characterizing the inherent noise~\cite{zeng2015distant,lin2016neural,feng2017effective}. These approaches focus on enhancing noise-tolerance of models instead of reducing noises from the source, hence, still suffer from the effects of noises in some ways. Some work considered utilizing many other kinds of knowledge besides KB~\cite{han2016global,liu2017heterogeneous}, to enrich the supervision knowledge. However, such studies suffer from the conflicts brought by multiple supervisions~\cite{ratner2016data}, and hard to benefit the existing relation extraction/classification models.

\subsection{Relation Extraction/Classification}

Relation classification aims to classify the given relation mention to a pre-defined relation type. Deep neural networks have shown promising results, and the representative progress was made by Zeng et al.~\shortcite{zeng2014relation}. To encode both past and future context information, Zhang and Wang~\shortcite{zhang2015relation} employed a bidirectional Recurrent Neural Network (Bi-RNN). To address the long-distance problem, some approaches based on Long Short-Term Memory networks (LSTM) have been proposed~\cite{zhang2015bidirectional,xu2015classifying}. Recently, Zhou~\shortcite{zhou2016attention} combined the attention model and bidirectional LSTM, achieving a significant improvements for relation classification.

Relation extraction can be regard as a pipeline of two separated tasks, i.e., named entity recognition and relation extraction. However, some studies consider extracting entities and relations in a single model. Most of these methods are feature-based ~\cite{ren2017cotype,yang2013joint,miwa2014modeling,li2014incremental}. Recently, Miwa and Bansal~\cite{miwa2016end} used a LSTM-based model to reduce such manual features. Zheng~\cite{zheng2017joint} converted the relation extraction to a sequence tagging problem, and proposed a LSTM-based encoder-decoder model to extract the entities and relations jointly without other redundant information, leading to the best results on the public dataset.

\subsection{Curriculum Learning}
The main idea of curriculum learning~\cite{bengio2009curriculum} is starting with the easiest aspect of a task and leveling up the difficulty gradually. Curriculum learning is mainly applied to solve various vision problems of Computer Vision (CV), such as tracking~\cite{supancic2013self}, face detection~\cite{lin2018active}, object detection~\cite{chen2015webly}, video detection~\cite{jiang2014self}, etc. Luo et al.~\shortcite{luo2017learning} applied curriculum learning to the task of relation classification. However, they used curriculum learning to address the cold-start of their model training, on a special dataset with explicit prior knowledge of data quality, which was different from our work.

\section{Conclusion}
In this paper, to alleviate the noise issue in distant supervision (DS), we take a new factor \textit{\textbf{time}} into consideration and propose a novel time-aware distant supervision (\textsc{Time-DS}). To make the most of \textit{\textbf{time}}, we consider two strategies, i.e., hard filter and curriculum learning. \textsc{Time-DS} benefits from these two strategies thus can guide the training process and further achieves better models on relation extraction/classification. The experimental results show the effectiveness of the time series \textsc{instance-popularity} and significant improvements on relation extraction/classification via feeding models into \textsc{Time-DS}. 

\bibliographystyle{ACM-Reference-Format}
\bibliography{my}


\end{document}